\documentclass{article}

\usepackage{microtype}
\usepackage{subfigure}
\usepackage{booktabs}
\usepackage{graphicx}
\usepackage{mwe}

\usepackage{hyperref}

\usepackage[accepted]{icml2024}

\usepackage{amsmath}
\usepackage{amssymb}
\usepackage{mathtools}
\usepackage{amsthm}

\usepackage[capitalize,noabbrev]{cleveref}

\usepackage[textsize=tiny]{todonotes}

\begin{document}

\twocolumn[
\icmltitle{Unlocking Fair Use in the Generative AI Supply Chain: A Systematized Literature Review}]




\begin{icmlauthorlist}
\icmlauthor{Amruta Mahuli}{MPI-SP}
\icmlauthor{Asia Biega}{MPI-SP}
\end{icmlauthorlist}

\icmlaffiliation{MPI-SP} {Max Planck Institute for Security and Privacy, Bochum, Germany}

\icmlcorrespondingauthor{Amruta Mahuli}{amruta.mahuli@mpi-sp.org }
\icmlcorrespondingauthor{Asia Biega}{asia.biega@mpi-sp.org}




\printAffiliationsAndNotice{}  

\begin{abstract}
Through a systematization of generative AI (GenAI) stakeholder goals and expectations, this work seeks to uncover what value different stakeholders see in their contributions to the GenAI supply line. This valuation enables us to understand whether fair use advocated by GenAI companies to train model progresses the copyright law objective of promoting science and arts. While assessing the validity and efficacy of the fair use argument, we uncover research gaps and potential avenues for future works for researchers and policymakers to address. 
\end{abstract}

\section{Introduction}
One of the demands made and accepted on account of the Writers’ Guild of America (WGA) protests\footnote{Art. 72 of Memorandum of Agreement for the 2023 WGA Theatrical and Television Basic Agreement: \url{https://www.wga.org/uploadedfiles/contracts/2023_mba_moa.pdf}} included disregarding generative artificial intelligence (GenAI) outputs as source material while determining writing credits and compensation. This demand highlighting the importance of credit and compensation for creative content creators was also voiced by Mr. Ashley Irwin during his congressional testimony\footnote{House holds hearing to examine the intersection of generative AI and copyright law — 05/17/23. Retrieved June 3, 2024 from \url{https://www.youtube.com/watch?v=isTgXmzoaXc}}. He reasoned that Art 1 section 8(8) of the US Constitution\footnote{U.S. Constitution. Art. I, § 8(8): \url{https://constitution.congress.gov/browse/essay/artI-S8-C8-1/ALDE_00013060/}} granted rights to authors and inventors to promote Science and arts, thus calling for policies that would regulate AI companies to generate their model using the ‘3Cs: consent, credit and compensation’ for creative workers (creatives). 

On the other hand, GenAI companies like OpenAI\footnote{OpenAI. 2019. Comment by OpenAI Regarding Intellectual Property Protection for Artificial Intelligence Innovation. Department of Commerce, United States Patent and Trademark Office. Retrieved December 22, 2023 from \url{https://www.uspto.gov/sites/default/files/documents/OpenAI_RFC-84-FR-58141.pdf}} are advocating for training models with copyrighted material as fair use based on how ‘transformative’ GenAI outputs will be, relying on Campbell v. Acuff-Rose Music\footnote{Campbell v. Acuff-Rose Music, 510 U.S. 569 (1994)} where the Supreme Court averred that “...the goal of copyright, to promote science and the arts, is generally furthered by the creation of transformative works.” 

Both arguments stem from the goal and justification for intellectual property law: to promote science and arts~\cite{Leaffer}. Ideally, these justifications form two sides of the same coin. However, in this case, and many lawsuits filed by creatives, they stand on opposing ends\footnote{Andersen et al v. Stability AI Ltd. et al, Docket No. 3:23-cv-00201 (N.D. Cal. Jan 13, 2023): ~\url{https://docs.justia.com/cases/federal/district-courts/california/candce/3:2023cv00201/407208/67}}\footnote{Main Sequence, Ltd. et al v. Dudesy, LLC et al, Docket No. 2:2024-cv-00711(Cal. Jan 25, 2024): ~\url{https://s3.documentcloud.org/documents/24377081/carlin-lawsuit.pdf}}\footnote{Tatjana Paterno and Layna Deneen. 2024. AI Threats Emerge In Music Publishers’ Battle With Big Tech. Hollywood Reporter. Retrieved June 3, 2024 from \url{https://www.hollywoodreporter.com/business/business-news/ai-threats-music-publishers-big-tech-1235767692/}}\footnote{Blake Brittain and Blake Brittain. 2024. Microsoft, OpenAI hit with new lawsuit by authors over AI training. Reuters. Retrieved June 3, 2024, from \url{https://www.reuters.com/legal/microsoft-openai-hit-with-new-lawsuit-by-authors-over-ai-training-2024-01-05/}}\footnote{Winston Cho. 2024. Scarlett Johansson, OpenAI Voice: Lawsuit for Actors Threatened. Hollywood Reporter. Retrieved June 3, 2024 from \url{https://www.hollywoodreporter.com/business/business-news/scarlett-johansson-ai-legal-threat-1235905899/}}. On the one hand, the creatives’ argument grounds itself in Lockean rhetoric of one's moral right to reap the benefits of one's labor. Conversely, OpenAI employs utilitarian rhetoric that views copyright as an incentivization system to enhance public welfare and advancement\cite{Leaffer}. 

The challenge for the judiciary and the policymakers lies in how they ideologically interpret these copyright law arguments, given that they both hold ground. In addition to this challenge of interpreting cogent arguments, economically, in 2021, the core copyright industries in the US added a value of \$1.810 trillion, which was 7.76\% of the U.S. GDP~\cite{IIPA}. Additionally, employment in the total copyright industries was more than 9.6 million creative workers, representing almost 5\% of the total US employment~\cite{IIPA}, making balancing this thin line of authorship rights and advancing cultural and scientific progress in society even more difficult. 

Therefore, this systematization of existing user studies draws from the conceptual framework of value similarity to corroborate the objectives of copyright law and the fair use doctrine for training GenAI models with the attributes valued by the stakeholders in their contributions to the GenAI supply chain. This paper begins by providing an overview of US copyright law and the philosophical groundings of copyright law. This section also maps the GenAI supply chain with the relevant stakeholders across every stage of the GenAI supply chain. Section 3 describes the method used in this paper to enable the analysis in Section 4. Section 5 discusses the results of the study. With the hope of uncovering research gaps and new future research directions, Section 6 concludes by reviewing the findings and proposing immediate computational challenges for better protection and implementation of the existing intellectual property frameworks, particularly in the context of GenAI.

\section{Methods}
\subsection{Data Collection}
This systematic literature review follows the PRISM statement while synthesizing existing Human-Computer Interaction (HCI) literature in English within the last 10 years since 2014 from 01/01/2014 to 04/30/2024~\cite{PRISM}.

We began by mapping stakeholders to different stages of the ML supply chain by relying on the supply chain conceptualized by Lee et al.~\cite{Cooper} as a structuring framework. Given the complexity and possibilities of involved actors, Lee et al.~\cite{Cooper}, for the sake of structure and clarity, our mapping of the stakeholders is approximate and has been summarized in Table 1.
\begin{table}
    \begin{tabular}{|p{5cm}|p{2cm}|}
    \hline
     \textbf{Stage in GenAI supply chain}  & \textbf{Stakeholder} \\
      \hline
     Creation of Creative Works    & Creatives\\
      \hline
      Data Creation, Dataset Creation, and  Dataset Curation   & Dataset Curators \\
       \hline
      Model Creation, Model Pre-training,       Model fine-tuning, and Alignment   &  ML practitioners\\
       \hline
     Deployment    & ML practitioners and UX designers\\
      \hline
     Generation    & End users \\
      \hline
    \end{tabular}
    \caption{GenAI supply chain and Stakeholder mapping }
    \label{Tab}
\end{table} 

We experimented with different keyword combinations based on our review goals and research questions to identify search query combinations to retrieve relevant existing literature. Given that our research questions seek to gauge the elements of value in the stakeholders’ contributions and ownership perceptions, we generalize by using different synonyms of the identified search strings where the star symbol (*) represents a wild card character, for example, recruit* could mean recruited or recruitment or recruit. We yielded 5895 papers using the following 4 sets of search strings for our search:
\begin{itemize}
    \item {[\textit{Search across the paper for}: "writer*" OR “knowledge workers” OR author* OR storyteller OR novelist* OR poet* OR co-author* OR “coauthor*” OR “cocreator” OR “co-creator” OR “creator” OR “artist” OR “dataset curat*” OR “database curat*” OR “dataset creat*” OR “database creat*” OR “ML practitioner*” OR “developer” OR “software developer” OR “ML developer” OR “computer scientist*” OR “data scientist*” OR “UX practitioner*” OR “UX designer*” OR “user*”] AND [\textit{Search within the abstract for}: "user studies" OR "survey" OR "interview*" OR "recruit*"] AND [\textit{Search across the paper for}: "goals" OR "expectation*" OR "own*" OR "incentiv*" OR "motiv*"]}
    \item {[[\textit{Search across the paper for}: "designer"] AND [\textit{Search with the abstract for}: "user studies" OR "survey" OR "interview*"] OR "recruit*"] AND [\textit{Search across the paper for}: "goals" OR "expectation*" OR "own*" OR "incentiv*" OR "motiv*"] AND [\textit{Search across the paper for}: "creativ*" OR "ux"]]}
    \item {[[\textit{Search across the paper for}: “GenAI" OR "generative artificial intelligence" OR "generative AI" OR "generative output"] AND [\textit{Search with the abstract for}: "user studies" OR "survey" OR  "interview*"] OR "recruit*"] AND [\textit{Search across the paper for}: "intellectual property" OR "copyright" OR  "patent"] OR "law"]]}
    \item {[[\textit{Search across the paper for}: "stakeholder*"] AND [\textit{Search within abstract for}: "user studies" OR "survey" OR "interview*" OR "recruit*"] AND [\textit{Search across the paper for}: "goals" OR "expectation*" OR "own*" OR "incentiv*" OR "motiv*"] AND [\textit{Search across the paper for}: "machine learning" OR "ML" OR "AI"] OR "artificial intelligence"]]}
\end{itemize}

To be eligible for our study, the publications met the following criteria:
\begin{itemize}
    \item Articles must be peer-reviewed full papers that were published over the last 10 years, limiting the results to the first 100 searches. Given the dynamic and fast-paced innovation in the machine learning and artificial intelligence domain, we believe that reviewing peer-reviewed full papers from the last 10 years will allow us to synthesize an understanding of the stakeholder perspectives.
    \item The selected articles must be full research papers that used qualitative user interviews as a research method for their research. To gauge stakeholder perspective on subjective issues such as valuation and expectations, we believe that user interviews and survey-based studies and papers have the potential to act as surrogates of authority to gain information from stakeholders. Therefore, we did not include full papers based on quantitative methods, non-user studies papers, posters, doctoral theses, books, book chapters, panel discussion reports, abstracts, or work-in-progress papers.
    \item The papers must relate to user perspectives on the thought process or workflow processes for generating works of authorship or generative artificial intelligence. Alternatively, we include papers relating to stakeholders’ perceptions of ownership regarding their contributions.
\end{itemize}

\begin{figure}
\centering
    \includegraphics[width=0.5\linewidth, height = 0.5\linewidth]{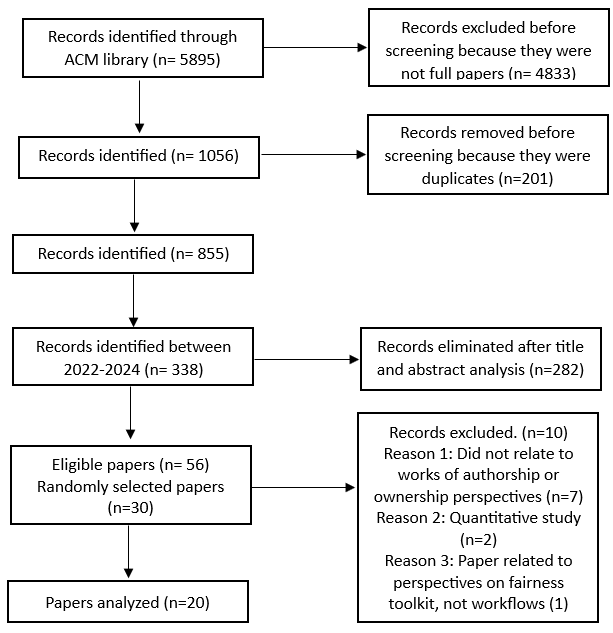}
    \caption{PRISMA flowchart of the article selection process}
    \label{Figure 1}
\end{figure}

Based on these criteria, out of the 5895 yielded papers, papers that did not meet the criteria, duplicates, and full papers that could not be found were removed to result in 855 papers. Given that this paper is a scoping study for the systematic literature review of these 855 papers, it restricts its scope. It systematically reviews papers from the past 3 years, from 2022 to April 2024, resulting in 338 papers.  A more thorough manual elimination process using title and abstract analysis was conducted to ensure that only the most relevant papers were reviewed. This round of elimination resulted in 56 papers selected papers. 31 papers were randomly selected for this systematic scoping literature review. After reviewing the selected papers, 10 were eliminated as they did not meet the inclusion/exclusion criteria mentioned in this section, resulting in 20 full papers forming the final dataset.

\subsection{Data Analysis}
Our units of analysis were the first 100 peer-reviewed, published full papers that came through the search as mentioned earlier queries, inclusion, and exclusion criteria from the ACM Guide to Computing literature library. We then analyzed the final filtered papers using MaxQDA 24, a qualitative and quantitative data analysis software. We used descriptive and in-vivo coding techniques to code the filtered papers for the following research questions: What do stakeholders in the GenAI supply think they value in their contributions to the GenAI supply chain?
We used affective methods such as in-vivo coding and value coding\cite{Saldana} by assigning codes to attributes that interviewed stakeholders, in their interview responses, mention or specify as incentivized or are considered ‘good’. The assigned codes were then analyzed using versus coding as an inspiration with labor theory and utilitarian theory as dichotomous groups.
Although we paid particular attention to the findings/results sections of the filtered papers to capture the interviewed stakeholders’ first-hand perceptions, we also carefully read through all the sections where stakeholders or the authors of the filtered papers mentioned any valuation or ownership perceptions.

\section{Results}
Using value coding and in-vivo coding, we identified over 600 codes, structured into 8 groups. The grounding for each theme with each stakeholder group is illustrated in Fig. 2:

It must be noted that some of the identified themes do not correspond directly to our research questions. However, it is indicative of a successful thematic analysis that all the identified themes do not simply correspond to the research questions~\cite{Vimpari, Braun}.  We have not collected and analyzed any descriptive statistics or metadata of the selected reviewed papers other than those required in the search terms, and inclusion and exclusion criteria.

\begin{figure}
\centering
    \includegraphics[width=0.85\linewidth, height=0.85\linewidth]{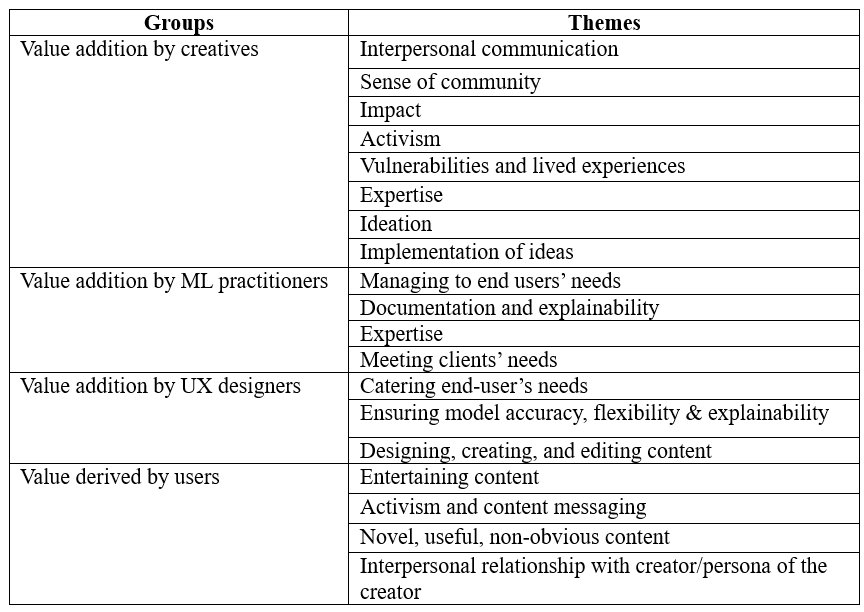}
    \caption{Mapping research questions with codes and themes in the paper}
    \label{Table 2}
\end{figure}

\textit{What do stakeholders in the GenAI supply think they value in their contributions to the GenAI supply chain?}
\subsection{Value added by creatives while creating expressive work}
\textit{Interpersonal communication and sense of community:} 
Some creatives believe they add value when their expressive work enables interpersonal communication and a sense of community. The text-to-image artist in one study~\cite{Chang} claimed that discovery (of new concepts) and a sense of community is rewarding for them as an artist: \begin{quote} “Everything is more fun when you can share it.”\end{quote} This emphasis on communicating ideas is also reflected in other studies\cite{Vimpari} emphasizing the human need to communicate and share ideas: \begin{quote} “We are humans. We want to communicate others how we feel, what ideas we have and that’s what it is. It’s an accelerator for human ideation and being able to reach.”\end{quote} In addition, participants in one study~\cite{Ma} highlight that their expressive work is a medium to \begin{quote}“maintain closeness with their audiences across platforms”.\end{quote} 

\textit{Impact and activism}: 
Some creatives believe that the value of expressive work lies in the impact and activism it inspires in the artists and the public. One of the participants in the study\cite{Chang}  noted that creating expressive work is \begin{quote} “not just having fun ... but [actually] making something that has some power.”\end{quote} A participant from another study\cite{Simpson} interviewing creatives on TikTok stated, \begin{quote}“And then January 6, happened. I was like, I understand this intricately. As someone who has studied white nationalism–as someone who has studied insurrections and fascism. Part of what I want to do is educate people to keep themselves safe. And every now and then I post book updates, every now and then I post [anonymized] content. But now it’s mostly news. It’s mostly talking about data transparency; it’s mostly talking about keeping people safe from threats they didn’t even know existed.” \end{quote}

\textit{Vulnerabilities and lived experience:} 
Interviews in studies show that creatives add value by tapping into their vulnerabilities and lived experiences to create expressive works. Simpson et al\cite{Simpson}, in their paper, noted that  'scripts, films, and produces videos on TikTok about surviving suicide, the intention behind creating these videos is not to receive attention and acclaim, but to help cope with their grief and help others cope with theirs.' A creative in the same study\cite{Simpson} noted that “just being yourself” garners more success. A writer interviewed for a paper\cite{Gero} to understand feedback networks emphasized the importance of the lived experience of their peers as an important factor in dictating whom the writer would ask for feedback: 
\begin{quote} "Since my brother is also Indian if I want to know how something reads to another Indian person, I will show him. But then if I’m writing a story about girlhood, I’ll send it to my friend Jen, who also writes about girlhood."
\end{quote} Another text-to-image artist in the same study\cite{Gero} noted that \begin{quote}“Art doesn’t live in a vacuum, nobody starts from scratch, everything is based on something else.”\end{quote} 

\textit{Expertise, ideation, and implementation of creative ideas}: 
Expertise, ideation, and implementation of creative ideas bring value to creatives' contributions while creating expressive content. Participants in a study\cite{Vimpari} consider the artist’s background and education relevant when using text-to-image generation in the creative process. They state that: \begin{quote} “artistic intention is required to guide the interaction, prompt engineering benefits from knowledge of art history styles and artistic techniques, and assessing the system’s output requires an intuitive understanding of aesthetics and rules to achieve e.g. color harmony.” \end{quote} A virtual influencer interviewed in a study\cite{Choudhry} noted that being a virtual influencer is more accessible because  \begin{quote}“you don’t have to be pretty to be a Virtual Influencer, just smart and be a good storyteller.” \end{quote} Another artist in a study\cite{Yan} shared: \begin{quote} "As a professional artist, shading, lighting, and the process after flatting is an eminent way to express my uniqueness. If I see the new features [that automate shadowing and lighting], I may feel an occupational crisis"\end{quote}  thus, emphasizing the importance of skill and creative imagination. A creative interviewed in a study focussed on multi-platform content creation\cite{Ma} stated that he \begin{quote} “understood that his audiences across platforms have different interests, so he curated different granularity of gaming topics of video content for each of his channels”\end{quote} showcasing his niche by curating and implementing ideas into different kind of content for his many channels.

\subsection{Value added by ML practitioners while developing ML models}
\textit{Managing end-users needs: }
ML practitioners in a study\cite{Hartikainen} shared that \begin{quote} "user needs to be the basis of the entire design.”\end{quote}  In another study\cite{Rahman}, participants corroborated this requirement when they stated that \begin{quote} “the success of the deployment does not include satisfying the functional or performance requirements but also how it is satisfying the target users.” \end{quote} However, in the former study\cite{Hartikainen}, some participants expressed their trust in their abilities to map user needs, and some participants noted that information about the user and their needs is often collected by conversing with the client and not the users. This paper\cite{Hartikainen} also stated that in 9 out of the 12 companies, the technical AI development team maps users' needs. In the remaining 3 companies, UX designers were responsible for mapping user needs.

\textit{Documentation and explainability:} 
Only one ML practitioner in the reviewed literature\cite{Laato} shared that they also need to “version each model and be able to connect them to datasets they were trained with”.

\textit{Expertise and ability to meet clients’ needs:} 
ML practitioners add value to the ML supply chain through expert domain knowledge and their ability to meet clients’ needs. A participant in a study\cite{Hartikainen} noted that \begin{quote} “In addition to the AI development, AI teams are responsible for the need assessment, early-phase interaction design, and communications with the client.”
\end{quote} Although companies often either built their models in-house or used third-party models\cite{Hartikainen}, model development, itself, was considered a highly iterative process where 'a lot of time is spent on tweaking the model parameters to achieve best possible performance.'\cite{Laato} 8 out of 12 participants in a study\cite{Hartikainen} noted that they do not have established processes when designing a model. Due to AI's uncertainty and dynamic nature, the ways of working strongly depend on the data, product, used AI technique, and the client\cite{Hartikainen}.  A participant in another study\cite{Laato} also corroborates this uncertainty when they expressed: "When [the ML model is] multiplying enormous matrices together and then applying some non-linear functions to them a hundred times in a row, how can anyone ultimately know what is happening there?" 

\subsection{Value added by UX practitioner while developing ML models’ interface}
\textit{Catering to end-users needs:}
Interviews with some ML practitioners in the reviewed study\cite{Hartikainen} indicate that UX practitioners are responsible for managing users’ needs: an ML practitioner noted, \begin{quote}“If we talk about the end-user, UI is the way to answer their needs. It’s hard for me to come up with any user-related practices during the early phase of development.”  
\end{quote} These needs include ensuring simplicity, ease of use, and easy-to-test interface for AI models\cite{Vimpari}. A UX practitioner in a study\cite{Feng} highlighted the importance of simplicity and ease of testing while designing the proof of concept for an interface, he noted: \begin{quote} “We probably want to start with something minimal that we can actually test with users.”\end{quote} 

\textit{Accuracy, flexibility, and explainability of ML models:}
The weight of user needs to ensure the accuracy, flexibility, and explainability of ML models falls upon the UX practitioners. A survey\cite{Feng} studying the collaborative practices and tools of UX practitioners noted that accuracy and flexibility were two common goals for UX practitioners. When comparing the workflow for designing AI and non-AI users, the participants in the same study\cite{Feng} noted that although the workflow for both user interfaces was not that different, they had to design to ensure user consent and awareness: \begin{quote} “When pure UX is concerned, I don’t see it as significantly different from any other user experience. I think that the user consent and awareness [of AI being used] needs to be a little bit more, but apart from that, in terms of experience, it needs to be like anything else I guess.” \end{quote}

\textit{Designing, creating, and editing content:}
Similar to creatives, UX practitioners add value to the ML supply chain by designing, creating, and editing content by applying their domain knowledge and expertise. UX practitioner participants in a study on text-to-image AI use\cite{Vimpari} noted that GenAI allows them to side-step their artists when they are too busy, unavailable, or when their engagement is too time-consuming: \begin{quote} "We definitely can use just straight-out stuff from Midjourney and give them to an artist instead and they’ll be super happy now. This is a good idea and they can, like copy, paste and basically paint over stuff then maturing, and then we can use that already."\end{quote}.  In addition to creating and editing content, interviewees in a study\cite{Khemani} described several other considerations while designing an interface that highlights and applies their expertise which includes \begin{quote} “how the language of the VUI is ‘built’ i.e., its ‘components’, how content is delivered, and how temporality (e.g., length of speech) shapes design (P8). Interviewees argued that language components such as syntax, grammatical rules, semantics, and lexical morphology dramatically influence the user experience of voice interfaces (P6) and therefore provided a go-to set of reflective considerations during VUI design.”\end{quote}  

\subsection{Value derived by users from GenAI outputs:}
\textit{Content that yields joy and entertainment:}
Users interviewed in the reviewed studies noted that they value expressive content that yields enjoyment. A participant in a study\cite{Li} on blind visual art patrons noted, \begin{quote} “Exploring art is simple to me, I feel relaxed and connected to the environment while I am in art galleries. It is very similar to watching a movie or playing a game...”   
\end{quote} For some patrons in the same study\cite{Li}, \begin{quote} “The goal for me to enjoy art is to understand the story and cultural background of different art pieces from different periods. Paintings and pictures are very language independent that you do not need to understand the language that the artists speak to appreciate their paintings..”\end{quote}   

\textit{Activism and messaging of expressive content:} 
In addition to enjoyment, participants in another study\cite{Choudhry} enjoyed expressive content for the activism and message of the expressive content. It must be noted that this study sought to understand the relationship between end-users and virtual influencers where some end-users stated that they followed the virtual influencer for \begin{quote} “the message they give, the art style, and mostly the stories they’re communicating." \end{quote} Another participant in the same study\cite{Choudhry} liked the positive message on environmental sustainability that the virtual influencers created. Some participants in this study benefitted from the endorsement that the virtual influencers advertised noting \begin{quote} "I’ll also pay more attention to the endorsement and likely form an opinion about the endorsing company that this must be a ’cool’ and ’in touch’ company for having an awareness of this.”\end{quote}     

\textit{Content that is novel, useful, and non-obvious:} 
Interviewees in a study noted that the participants characterized good ideas as having at least two of the properties of novelty, usefulness, or nonobviousness (but not always all three)\cite{Inie}.  Participants in another study\cite{Choudhry} surveying virtual influencer followers validated the importance of novelty and nonobviousness: \begin{quote} “The limitations that traditional influencers face is simply the fact that they are human. As an actual living being, there are certain boundaries to the things they can do. On the other hand, a virtual influencer is just the creation of a human. It can and will do anything one can dream of. This makes their content unique, be it Reels, posts, or stories."\end{quote}. Another participant in the same study\cite{Choudhry} noted, \begin{quote} “Maybe because I know they are fake, and someone is working really hard to make a fake person??? It’s like the "artist" or "creator" or whatever of the virtual influencer is showcasing their abilities/progression. Like an artist would, posting their art.” 
\end{quote}; A participant in the same study\cite{Choudhry} stated \begin{quote} “Virtual influencers are liked for their aesthetic. Being virtual is sort of part of their thing.”\end{quote}

\textit{Connecting with the persona and interpersonal relationship with the creator:} 
While being virtual adds value to the creativity of virtual influencers, it also causes some of the participants to be emotionally detached from them, thus enabling them to engage more confidently with the influencers\cite{Choudhry}. One participant in a study\cite{Choudhry} noted that \begin{quote} “With a VI account, I comment more easily because it’s lighter, I never comment on other accounts because I figure that stars or other celebrities can read and I don’t necessarily seek more interaction with them. With a VI account, I don’t really think about the person behind the account because I figure the person is more detached, for bee influencing for example it’s just a small part of their job I guess.”\end{quote} This detachment and sense of anonymity, however, renders a greater responsibility on the creatives as users believe\cite{Choudhry} that such creatives can voice their opinions more freely and engage on more sensitive issues because they are not human and therefore do not have to worry about repercussions such as public hate: \begin{quote} “I think that, if she would live in our world and she is up to date about things happening in our society, she would probably give an unfiltered and pure opinion because she is not a real person. She will be able to tell things a normal person wouldn’t want to say. . .if you can use an AI to say things humans are going to get canceled for then why not. . . Social media is very toxic and people can be hated based on their opinion. Wouldn’t it be great to use AI for sensitive topics?”\end{quote}

\section{Analysis and Discussion}
\subsection{Does copyright law protect the attributes valued by stakeholders in GenAI supply chain?}
\begin{figure}
    \centering
    \includegraphics[width=1.00\linewidth, height=1.00\linewidth]{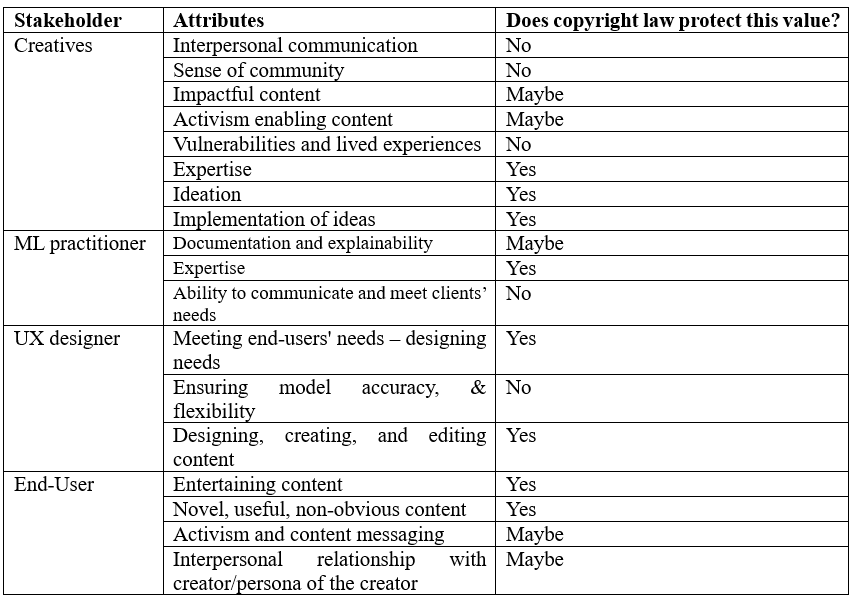}
    \caption{Mapping copyright law values with valued attributes of stakeholders}
    \label{Figure 3}
\end{figure}

For our analysis, we rely on the concept of value similarity introduced by \cite{Mehrotra, Cruciani}. In \cite{Cruciani} paper, a simulation experiment imitating the value similarity model proved that similarity in values or attributes can be a driving force for cooperation and trust in the regulation and design of public policy. Given this context, we attempt to understand the value similarity between the values behind exclusive rights enshrined by copyright law and the attributes valued by stakeholders in the GenAI system, a summary of which can be found in Table. 

Creatives’ value addition in terms of expertise, novelty in ideation, and implementation of those ideas are protected under copyright law and are appreciated by end-users consuming the implementation of these ideas. Both creatives and end-users appreciate enjoyable, impactful content that promotes advocacy. Given the expressive nature of this type of content, it is protected under copyright law. 

Creatives’ value additions by doing work that fosters interpersonal communications with their fans/followers and a sense of community are also appreciated by end-users. However, these attributes are not protected by copyright law. Creatives also add value when they share their life experiences and vulnerabilities with the end-users. However, this kind of input is not protected by copyright law.

ML practitioners make a distinction between end-users' needs and clients’ needs. They add value to the ML supply chain when communicating and meeting clients’ needs. Their work also adds value to the supply chain through their expertise and ensuring adequate documentation and explainability of the model. Although ML practitioners’ expertise is protected through the models, they develop along with their work process of documenting and ensuring the explainability of the model through the document itself, their ability to communicate with the clients and meet the clients’ needs remains unprotected under copyright law.
The UX practitioners’ value addition in terms of meeting end-users' needs through the UX design and interface itself can be protected under copyright law, along with the contributions they make when they design, create, and edit content. However, their contribution when they ensure flexibility in the model accuracy and agility remains unprotected.  

\textit{Recommendations}: Given this gap between the values protected under copyright laws and the attributes valued by stakeholders, this paper, therefore, calls for the need to reinvent copyright law with computational solutions that go beyond the letter of the law but encapsulate the spirit of the law. In particular, given the analysis of these studied papers, as future work, researchers may consider developing tools that offer a communal environment for fostering interpersonal communication and a sense of community with other stakeholders in the ecosystem. Researchers may also consider creating tools that enhance the GenAI infrastructure by supporting creative autonomy and providing nuanced technical means of self-expression while taking into account stakeholder needs. 

\subsection{Does the advocacy of fair use doctrine enable the progress of copyright law’s objectives?}
The analyzed papers also show that ML practitioners believe that data processes such as training and fine-tuning models are the most challenging aspects of their workflow. With GenAI companies like OpenAI advocating for training models as fair use and reports suggesting that researchers may run out of data to train, it is pertinent to ask whether using the fair use doctrine to train GenAI models is, in fact, producing transformative content that is enhancing or even promoting the enhancement of public knowledge of science and arts.

OpenAI, in its submission to the Copyright Office \footnote{OpenAI. 2019. Comment by OpenAI Regarding Intellectual Property Protection for Artificial Intelligence Innovation. Department of Commerce, United States Patent and Trademark Office. Retrieved December 22, 2023 from \url{https://www.uspto.gov/sites/default/files/documents/OpenAI_RFC-84-FR-58141.pdf}}, clarified that AI systems learn patterns from the training corpus and use those patterns to generate ‘novel’ expressive works which ‘share some commonality with works in the corpus subject to the model learning from ‘enormous number of works’ . Furthermore, OpenAI argues that such infringement “is an unlikely accidental outcome.” However, a report by the US Congressional Research Service\footnote{Congressional Research Service. 2023. Generative Artificial Intelligence and Copyright Law. US Congress. \url{https://sgp.fas.org/crs/misc/LSB10922.pdf}}  brings forth a contradicting narrative by the Getty Images lawsuit \footnote{Getty Images (US) Inc \& Ors v Stability AI Ltd [2023] EWHC 3090 (Ch)} that alleges “Stable Diffusion at times produces images that are highly similar to and derivative of the Getty Images.” This report also brings attention to a paper\cite{Somepalli} that found “a significant amount of copying” in less than 2\% of the images created by Stable Diffusion where the authors claimed that their methodology “likely underestimates the true rate” of copying . While 2\% may seem like a significantly small number, and it can be agreed that generative AI does have the potential to create transformative work, extant literature\cite{Laato} indicates that ML practitioners who develop and train these models face difficulties predicting whether the outcomes of such AI models will be original and transformative for society at all: \begin{quote} “When [the ML model is] multiplying enormous matrices together and then applying some non-linear functions to them a hundred times in a row, how can anyone ultimately know what is happening there?”\end{quote} 

These characteristics of ML models being black boxes, their unpredictability, AI mimicry, and overfitting are also impacting the drive of creatives to create expressive works. In a survey of more than 1000 artists\cite{Shan} where more than 95\% of the artists post their art online, 53\% of them anticipate reducing or removing their online artwork if they have not done so already. Out of these artists, 55\% of them believe reducing their online presence will significantly impact their careers. One participant stated, \begin{quote} “AI art has unmotivated myself from uploading more art and made me think about all the years I spent learning art.”\end{quote} 78\% of artists anticipate AI mimicry would impact their job security, and this percentage increases to 94\% for the job security of newer artists. Further, 24\% of artists believe AI art has already impacted their job security, and an additional 53\% expect to be affected within the next 3 years.  This impact on creatives' jobs is also seen in the accepted demands of the WGA which highlights the precarious artistic and financial security in the creative industry due to the advent of generative AI models. 

The argument of fair use falls flat if we apply Leaffer's considerations for fair use doctrine\cite{Leaffer}  which include a) that the use must not impose economic harm to the original copyright owners and b) that the use is one that provides something new and potentially valuable to society. Firstly, GenAI, in its present version, has raised significant apprehensions amongst the original creatives content creators regarding the economic harms they face due to GenAI outputs.  Secondly, it is evident from the extant literature that although GenAI has the potential to create transformative work, as of the submission of this paper, GenAI is still plagued with issues of AI mimicry, overfitting, and excess memorization. 

\textit{Recommendations}: Therefore, unless these problems of AI mimicry, overfitting and excess memorization are resolved computationally, GenAI will continue to create pattern-based content that is neither original nor novel, failing to validate its claim for eligibility for the fair use exception.

\subsection{Conclusion}
This systematic literature review reveals attributes that stakeholders in the ML supply chain believe are incentivized in their contributions. Although GenAI models and their outputs find their origins for intellectual property ownership in the contributions of the data curators, the lack of studies that focus on data curators’ perspectives in this regard highlights a gap in the existing literature. This paper also uncovers computational gaps that expose cracks in the GenAI companies' call for training data for GenAI models to be considered fair use. It also recognizes the need to reinvent and complement copyright law with computational solutions for better implementation of the letter and spirit of the law. It, therefore, spells out computational problems in GenAI models and outputs for researchers and practitioners that demand immediate resolution for better implementation and enforcement of intellectual property laws and governing ownership and stakeholder value beyond the protections afforded by copyright.
\bibliography{main}
\bibliographystyle{icml2024}



\end{document}